\documentclass{article}

\PassOptionsToPackage{numbers, compress}{natbib}

\usepackage[sglblindworkshop, preprint]{neurips_2026}
\workshoptitle{Machine Learning for Systems}

\usepackage[utf8]{inputenc}
\usepackage[T1]{fontenc}
\usepackage{hyperref}
\usepackage{url}
\usepackage{booktabs}
\usepackage{amsfonts}
\usepackage{nicefrac}
\usepackage{microtype}
\usepackage{xcolor}

\title{When Agents Implement Systems: A Case Study in Defects, Detection, and Evaluation Rigor}

\author{%
  Phanindra Reddy Madduru \\
  Amazon.com \\
  \texttt{maddurup@amazon.com} \\
}

\begin{document}

\maketitle

\begin{abstract}
As LLM coding agents increasingly perform end-to-end engineering work rather
than single-function code completion, we lack empirical characterization of
how they behave when implementing genuinely \emph{systems}-level
requirements: a storage schema, concurrency and async orchestration,
cross-process configuration correctness, and a retrieval-filtering
precision trade-off. We present a case study of an LLM
coding agent implementing a multi-component data system (a persistent graph
store, a filtered vector index, an async ingestion pipeline, and a
real-time visualization client) against a detailed pre-existing
specification --- storage technologies, schema, entity-resolution
algorithm, and retrieval-filtering strategy fixed in advance; the agent's
autonomy was in the implementation, in diagnosing and fixing defects it
introduced while building to that spec, and in interaction-design choices
the spec left open. Over a single
extended session, we catalog five such defects, categorized by which
systems constraint each violates and by whether it was caught by automated
testing or only by empirical probing (rendered screenshots, timing
measurements) with no type-level or static signature. We further evaluate,
on a public multi-hop QA benchmark (HotpotQA, distractor setting), the one
retrieval trade-off specified in that architecture --- restricting
candidates to a graph-identified entity set before ranking, versus
unfiltered semantic search --- reusing the benchmark's gold evidence labels
as a stand-in for entity identification (which we did not have LLM access
to run in this session) and for a standard information-retrieval recall
metric rather than the benchmark's own answer-accuracy metrics. Across
$k \in \{1,3,5,10\}$ and $n=100$ questions against a 2{,}994-paragraph
pooled corpus, filtered recall reaches 1.0 by $k=3$ --- expected once
candidates are restricted to the gold-titled paragraphs themselves ---
while unfiltered search recovers all required evidence only 69\% of the
time even at $k=10$, a gap that holds at every $k$ (sign test,
$p<10^{-4}$). We close with a candid discussion of where the agent's
autonomy succeeded versus required human correction, including one
instance where a claimed performance fix was never re-measured on the
regression that motivated it --- an unverified engineering claim that the
self-reported development loop, in this instance, had no mechanism to
catch.
\end{abstract}

\section{Introduction}

LLM coding agents are increasingly deployed not for isolated code
completion but for open-ended, multi-file engineering work: implementing
data models, wiring asynchronous job queues, and debugging failures that
only manifest at runtime. Existing evaluations of agentic coding largely
score whether an agent resolves a pre-specified, bounded issue in an
existing codebase \citep{jimenez2024swebench}, and existing agent
architectures for combining reasoning with tool use (retrieval, code
execution) are evaluated on task success rate \citep{yao2023react}.
Neither setting captures a different failure mode: an agent given a
detailed specification for a multi-component system, building it end to
end across many files and pipeline stages, and introducing --- then having
to diagnose and fix --- its own systems-level defects along the way. This
shifts the relevant evaluation question from ``does the generated function
pass its unit test'' to ``does the agent behave reliably when implementing
systems engineering under the constraints systems engineers actually
face'' --- consistency across a schema shared by multiple call sites,
correctness of idempotent storage operations under repeated execution,
reliability of configuration that must resolve identically regardless of
which process launches it, and precision trade-offs in retrieval
filtering. ML-driven systems design has an established track
record when the target is a well-specified optimization problem, such as
learned chip placement \citep{mirhoseini2021chip}; our case study instead
concerns implementation reliability against a specification that is itself
fixed and detailed --- whether an agent introduces, and can self-correct,
systems-level defects while building to it.

We report an observational case study: a single LLM coding agent (Claude,
operated via a CLI coding tool) implemented a multi-component data system
end to end --- a persistent, schema-driven graph store; a
metadata-filtered vector index; an asynchronous multi-stage ingestion
pipeline; and a force-directed graph visualization client --- against a
detailed pre-existing specification (Section~2). Over the course of the
session, five defects were introduced by the agent's own implementation
work and later diagnosed and fixed, either by the agent proactively or in response to a
user-reported symptom. We treat this transcript as data: we categorize
each defect by which systems constraint it violates --- reliability,
correctness, consistency, or efficiency --- and by how it was actually
caught (automated test vs.\ empirical/visual probing). We then evaluate
one specified engineering trade-off in that architecture --- filtering
retrieval candidates by a graph-identified entity set before ranking,
versus unfiltered semantic search --- quantitatively on a public external
benchmark, rather than the internal corpus the case-study system was
originally developed against.

We are explicit about what this paper is not: graph-guided retrieval
filtering is not new --- it is established in the retrieval-augmented
generation and GraphRAG literature \citep{lewis2020rag,edge2024graphrag}.
The contribution is Section~3's defect catalog; Section~4's benchmark is
one reproducible illustration of a specified trade-off, not a new
retrieval technique.

Contributions: (1) a catalog of five systems defects introduced during
spec-directed implementation and later diagnosed and fixed by the
implementing agent, categorized by constraint type and detection method;
(2) an externally-benchmarked evaluation, with a significance test and a
$k$-sweep, of one specified retrieval-filtering trade-off in that
architecture; (3) a candid discussion of an evaluation-rigor gap the agent
itself exhibited
--- claiming a fix without re-measuring the regression it targeted --- and
what it implies for trusting an autonomous agent's self-reported
engineering claims.

\section{Case Study System and Methodology}

The target system is a multi-component data pipeline with four parts
relevant to systems evaluation, independent of its application domain:
(i) a graph store (embedded relational database, full-text fuzzy lookup, a
recursive multi-hop traversal query, write-ahead-log concurrency), in the
same graph-based-system family as prior heterogeneous-graph
deployments \citep{madduru2023fraud}; (ii) a vector index with
metadata-based candidate filtering; (iii) an asynchronous ingestion
pipeline (multi-stage extract/resolve/embed/link, orchestrated over a task
queue) required to be idempotent under repeated execution, the same
ingestion-scalability concern addressed for graph databases more
broadly \citep{madduru2026edgevalue}; and (iv) a force-directed graph
rendering client, in the same graph-visualization space as prior
explainability tooling \citep{zhang2023explainer}.

A pre-existing design document fixed the storage technologies, schema,
entity-resolution algorithm, and retrieval-filtering strategy Section~4
evaluates; left to the agent were the implementation itself, the defects
it introduced (Section~3), and interaction-design details beyond the
mandated technology choices. The methodology for this paper is
observational: we log every defect with a directly attributable code diff
and observable symptom surfaced during development or subsequent use, the
diagnostic path that led to its root cause, and the fix, drawing only on
artifacts already produced during the session (code diffs, test output,
screenshots, timing) rather than a protocol decided in advance.

\section{Observed Defects, Categorized by Systems Constraint}

Table~\ref{tab:defects} summarizes five defects, spanning four systems
constraints; two instances share the correctness category but manifest at
different layers (storage-layer idempotency vs.\ render-time visual
output). We discuss the two most instructive below.

\begin{table}[h]
\centering
\small
\begin{tabular}{@{}p{2.1cm}p{4.6cm}p{2.6cm}p{3.4cm}@{}}
\toprule
\textbf{Constraint} & \textbf{Defect} & \textbf{Caught by} & \textbf{Fix locus} \\
\midrule
Reliability & Config path resolution depended on process launch directory; broke only for a non-default entrypoint & Runtime exception, one entrypoint only & Centralized, anchored to module location, not caller cwd \\
Correctness (storage) & Storage upsert keyed on randomly generated IDs; idempotency claim violated under re-execution & Repeated-execution test added after the fact & Deterministic IDs derived from stable inputs \\
Consistency & Graph traversal returned edges in both directions but only fetched endpoint nodes in one direction & Client-side runtime crash (rendering library) & Explicit endpoint backfill; recurred at a second call site \\
Efficiency & Sequential per-stage network calls; wall-clock cost scaled with input size with no concurrency & User-observed latency & Bounded concurrent execution \\
Correctness (rendering) & Screen-space label size computed with a lower bound that broke an intended scale-cancellation & Automated screenshot comparison; no static signature & Removed the erroneous bound \\
\bottomrule
\end{tabular}
\caption{Defects introduced and fixed during implementation, categorized by the constraint each violates and by how each was actually detected.}
\label{tab:defects}
\end{table}

\textbf{Consistency defect, recurring across call sites.} The graph
traversal component followed only outgoing edges when expanding a node's
neighborhood, while a separate function collecting edges for the same node
returned edges in both directions. The result: an edge could be returned
whose endpoint node was never included in the node set handed to a
downstream consumer. This was invisible to the type system and to the
existing test suite (which asserted on returned edges, not the joint
node/edge consistency invariant), manifesting only as a runtime crash in
an unrelated client --- a force-directed rendering library --- resolving a
link's endpoint missing from its node array. The same pattern was
independently reintroduced at a second call site (an HTTP endpoint
exposing the same operation) before the invariant was made explicit and
tested directly.

\textbf{Rendering correctness defect, caught only visually.} A canvas
label was sized as $\max(c, b/s)$ for on-screen constant $b$, current zoom
$s$, and floor $c$, intending scale-invariant text size. The floor broke
the intended cancellation: once $s$ exceeded $b/c$, the floored value was
re-multiplied by the zoom transform on render, producing screen-filling
text. No static check could catch this: it is a purely runtime, visual
effect with no failing assertion short of rendering a frame and measuring
pixels. It was caught only by an automated screenshot comparison during
otherwise unrelated interaction testing.

\section{A Specified Retrieval Trade-off: External Benchmark Evaluation}

The case-study architecture specifies one retrieval trade-off precisely
enough to benchmark externally: restrict vector-retrieval candidates to a
graph-identified entity set before ranking (\emph{filtered}), versus rank
the full corpus by embedding similarity alone (\emph{unfiltered}, the
standard RAG setting \citep{lewis2020rag}) --- filtering narrows the
search space using structural information, in the spirit of
graph-augmented retrieval \citep{edge2024graphrag}, at the cost of
depending on correct upstream entity identification.

We evaluate this decision on HotpotQA (distractor setting) \citep{yang2018hotpotqa},
a public multi-hop QA benchmark whose distractor paragraphs are
specifically constructed to be lexically similar to the gold evidence,
making it a natural stress test for this exact trade-off. \textbf{We
emphasize what this evaluation is not:} we do not reproduce HotpotQA's own
benchmark task or its published answer-EM/F1 and supporting-fact-EM/F1
metrics, and our numbers are not comparable to any leaderboard entry for
that dataset. We reuse only its raw paragraphs and gold supporting-fact
labels as ground truth for a narrower, standard information-retrieval
question: given a fixed corpus and a fixed embedder, does restricting
candidates to a known-relevant entity set change whether the retriever
surfaces the gold evidence paragraph(s)?

We pool the context paragraphs from 300 randomly sampled validation
questions (2{,}994 paragraphs total) into one shared corpus, using each
paragraph's Wikipedia article title as its entity identifier (no LLM
extraction required: HotpotQA paragraphs are single-entity by
construction). The first 100 of these 300 questions serve as our
evaluation set, so each evaluation question's distractor pressure comes
from the full 2{,}994-paragraph pool, not just its own 10 context
paragraphs. For each evaluation question we compare an \textbf{unfiltered}
condition (top-$k$ nearest paragraphs by cosine similarity over the entire
pool) against a \textbf{filtered} condition (candidates restricted to
paragraphs whose title is in the question's gold supporting-fact title
set --- a stand-in for a correctly functioning entity-identification stage
upstream --- then ranked by similarity within that restricted set).

We report recall@$k$ against HotpotQA's annotated supporting-fact
paragraphs for $k \in \{1,3,5,10\}$ (Table~\ref{tab:ablation}), the
fraction of questions where \emph{both} gold paragraphs are recovered
(exact match), and a two-sided exact sign test on paired per-question
recall differences. We did not have LLM access to run the full
entity-identification stage in this session; the gold-title substitution
isolates the retrieval-filtering decision itself, cleanly separated from
entity-identification quality, which we flag as the main threat to
generalizing these numbers to a fully automated pipeline.

\begin{table}[h]
\centering
\small
\begin{tabular}{lcccc}
\toprule
& $k=1$ & $k=3$ & $k=5$ & $k=10$ \\
\midrule
Mean recall@$k$, unfiltered            & 0.420 & 0.685 & 0.765 & 0.845 \\
Mean recall@$k$, filtered              & 0.500 & 1.000 & 1.000 & 1.000 \\
Exact match, unfiltered                & 0.000 & 0.410 & 0.540 & 0.690 \\
Exact match, filtered                  & 0.000 & 1.000 & 1.000 & 1.000 \\
Sign test $p$ (paired, per question)   & $3\times10^{-5}$ & $3\times10^{-18}$ & $3\times10^{-14}$ & $9\times10^{-10}$ \\
\bottomrule
\end{tabular}
\caption{Retrieval evidence recovery, $n=100$ HotpotQA validation questions, pooled corpus of 2{,}994 paragraphs (see text for interpretation).}
\label{tab:ablation}
\end{table}

Two results in Table~\ref{tab:ablation} are informative; the rest follow
near-tautologically from the design: the filtered candidate pool is
exactly the two gold-titled paragraphs, always present in the corpus by
construction, so any $k\geq2$ recovers them regardless of ranking quality.
The results \emph{not} guaranteed by construction are (a) filtered
recall@1 $=0.500$, where ranking within the 2-paragraph pool still
matters, and (b) the unfiltered condition's shortfall, which does not
close with a larger budget: exact match reaches only 69\% at $k=10$ ---
double the budget of the filtered condition's already-100\% $k=5$
result --- and every sign test rejects the null of no paired difference at
$p<10^{-4}$. This persistent unfiltered shortfall, not the filtered
ceiling, is the finding: topically similar but wrong-entity evidence
crowds out correct evidence under pure similarity ranking, and more
retrieval budget does not fix it.

\section{Discussion}

\textbf{Where autonomy succeeded vs.\ required correction.} In two of
five defects, the agent diagnosed the root cause from a single symptom and
applied a structural fix rather than a local patch --- centralizing
configuration-path resolution, and generalizing the endpoint-backfill fix
after recognizing the same pattern at a second call site. A related but
uncatalogued issue (outside Table~\ref{tab:defects}'s five) shows the
opposite pattern: the visualization client's initial design re-fetched and
re-rendered on every click; a human caught this only through interactive
testing, since each operation was locally correct and the failure was
emergent.

\textbf{An evaluation-rigor gap.} The concurrency fix in Table~\ref{tab:defects}
was motivated by a 242-second latency on an 18-chunk document. After
parallelizing the affected stage, the agent validated the fix only on a
smaller document, reporting it effective without re-measuring latency on
the \emph{same} input that motivated the fix --- conflating
``architecturally sound'' with ``verified on the regression it targets.''
\textbf{Limitations:} single-session, single-agent case study; no claim
that defect frequencies generalize --- including to smaller, more
efficient models, whose capability trade-offs differ from the
frontier-scale agent used here \citep{gupta2025slm}; the retrieval
ablation substitutes gold entity labels for query understanding, bounding
only the filtering decision in isolation.

\section{Conclusion}

Several of the five cataloged defects had no static or type-level
signature, caught only by empirical probing --- arguing for execution-based,
not just static, evaluation of agentic systems work. The isolated
retrieval-trade-off benchmark and the unverified-fix case point to the same
gap: nothing in this session's development loop forced re-measurement
against the regression a fix targeted, or against an external benchmark
rather than self-report. The ablation script and its results (Table~\ref{tab:ablation})
are retained alongside this paper's source so the retrieval-filtering
comparison can be independently re-run.

\bibliographystyle{plain}

\end{document}